\documentclass[letterpaper, 10 pt, conference]{ieeeconf}
\IEEEoverridecommandlockouts
\usepackage[colorlinks,bookmarksopen,bookmarksnumbered,citecolor=black,urlcolor=black]{hyperref}

\usepackage{booktabs,multirow}
\usepackage{algorithm}
\usepackage[noend]{algpseudocode}
\usepackage[english]{babel}
\usepackage[pdftex]{graphicx}
\usepackage{adjustbox}
\DeclareGraphicsExtensions{.pdf,.jpg,.jpeg,.png}

\usepackage{graphicx} 
\usepackage{color, colortbl}
\usepackage{bm}
\usepackage[caption=false]{subfig}
\usepackage{amsmath} 
\usepackage{amssymb}  
\usepackage{physics}
\usepackage[table]{xcolor}
\usepackage{algcompatible}

\algblockdefx{FORALLP}{ENDFAP}[1]%
  {\textbf{for all }#1 \textbf{do in parallel}}%
  {\textbf{end for}}

\usepackage{bm}
\usepackage{glossaries}
\glsdisablehyper
\hypersetup
{
	pdftitle = {Learning How to Walk: Warm-starting Optimal Control Solver\\with Memory of Motion},
	pdfauthor = {Teguh Santoso Lembono, Carlos Mastalli, Pierre Fernbach, Nicolas Mansard,  Sylvain Calinon},
	pdfsubject = {Submitted to ICRA},
  pdfkeywords = {memory of motion, optimal control, legged robotics},
	pdftoolbar = true,
	colorlinks = true,
	linkcolor = black,
	citecolor = black,
	urlcolor = black,
}
\newacronym{tsid}{TSID}{Task-Space Inverse Dynamics}
\newacronym{mpc}{MPC}{Model Predictive Control}
\newacronym{ddp}{DDP}{Differential Dynamic Programming}
\newacronym{fddp}{FDDP}{Feasibility-prone Differential Dynamic Programming}
\newacronym{dofs}{DoFs}{Degrees-of-Freedoms}

\makeatletter
\newcounter {subsubsubsection}[subsubsection]
\renewcommand\thesubsubsubsection{\thesubsubsection .\@arabic\c@subsubsubsection}
\newcommand\subsubsubsection{\@startsection{subsubsubsection}{4}{\z@}%
                                     {-3.25ex\@plus -1ex \@minus -.2ex}%
                                     {1.5ex \@plus .2ex}%
                                     {\normalfont\normalsize\bfseries}}
\newcommand*\l@subsubsubsection{\@dottedtocline{3}{10.0em}{4.1em}}
\newcommand*{\subsubsubsectionmark}[1]{}
\makeatother

\newcommand\copyrighttext{%
  \footnotesize \textcopyright 2020 IEEE. Personal use of this material is permitted.
  Permission from IEEE must be obtained for all other uses, in any current or future
  media, including reprinting/republishing this material for advertising or promotional
  purposes, creating new collective works, for resale or redistribution to servers or
  lists, or reuse of any copyrighted component of this work in other works.}
  
\newcommand\copyrightnotice{%
\begin{tikzpicture}[remember picture,overlay]
\node[anchor=south,yshift=10pt] at (current page.south) {\fbox{\parbox{\dimexpr\textwidth-\fboxsep-\fboxrule\relax}{\copyrighttext}}};
\end{tikzpicture}%
}

\graphicspath{{./figures/}} 

\usepackage{tikz}
\usetikzlibrary{shapes,arrows}
\tikzstyle{decision} = [diamond, draw, fill=blue!20, 
    text width=4.5em, text badly centered, node distance=3cm, inner sep=0pt]
\tikzstyle{block} = [rectangle, draw, fill=blue!20, 
    text width=10em, text centered, rounded corners, minimum height=4em]
\tikzstyle{line} = [draw, -latex']
\tikzstyle{cloud} = [draw, ellipse,fill=red!20, node distance=3cm,
    minimum height=2em]
\tikzstyle{data} = [draw,trapezium,trapezium left angle=70,trapezium right angle=-70,minimum height=1cm, align=center]

\definecolor{LightCyan}{rgb}{0.6,1,1}
\definecolor{darkgreen}{rgb}{0.0, 0.5, 0.13}

\title{Learning How to Walk: Warm-starting Optimal Control Solver\\with Memory of Motion}

\author{Teguh Santoso Lembono$^{1,2}$\quad Carlos Mastalli$^{3}$\quad Pierre Fernbach$^{3}$\quad Nicolas Mansard$^{3}$\quad  Sylvain Calinon$^{1,2}$
\thanks{$^1$ Idiap Research Institute, Martigny, Switzerland \quad $^2$ EPFL, Lausanne, Switzerland \quad $^3$ Gepetto Team, LAAS-CNRS, Toulouse, France.}
\thanks{This work was supported by the European Union under the EU H2020 collaborative project MEMMO (Memory of Motion, \url{http://www.memmo-project.eu/}), Grant Agreement No. 780684.}
}

\begin{document}

\maketitle

\copyrightnotice

\thispagestyle{empty}
\pagestyle{empty}

\begin{abstract}
In this paper, we propose a framework to build a memory of motion for warm-starting an optimal control solver for the locomotion task of a humanoid robot. We use HPP Loco3D, a versatile locomotion planner, to generate offline a set of dynamically consistent whole-body trajectory to be stored as the memory of motion. The learning problem is formulated as a regression problem to predict a single-step motion given the desired contact locations, which is used as a building block for producing multi-step motions. The predicted motion is then used as a warm-start for the fast optimal control solver Crocoddyl. We have shown that the approach manages to reduce the required number of iterations to reach the convergence from $\sim$9.5 to only $\sim$3.0 iterations for the single-step motion and from $\sim$6.2 to $\sim$4.5 iterations for the multi-step motion, while maintaining the solution's quality. 
\end{abstract}

\section{Introduction}
\label{sec:introduction}

Legged locomotion is often achieved by first computing a sequence of contacts, generating a stable centroidal trajectory, and finally computing a whole-body motion through inverse dynamics~\cite{winkler2015planning}\cite{carpentier2018multicontact}.
However, this approach cannot properly regulate angular momentum because (a) centroidal trajectory optimization~\cite{ponton2018time}\cite{aceituno_cabezas-ral17} does not consider the limbs momenta, and (b) whole-body control~\cite{del2016robustness}\cite{fahmi2019passive} is  instantaneous control action that theoretically cannot properly regulate the angular momentum since it represents a nonholonomic constraint on the multibody dynamics~\cite{wieber2006holonomy}. 


Recently, optimal control is getting more attention in legged robots due to (a) its ability to properly control angular momentum~\cite{budhiraja2018differential}, and (b) they can be deployed for real-time control~\cite{koenemann2015whole}\cite{neunert2017trajectory}.
In this vein, we have proposed an efficient multi-contact optimal control framework called Crocoddyl~\cite{mastalli2020}.
This framework relies on a novel multiple-shooting optimal control solver called \gls{fddp}.
We have shown that Crocoddyl can generate highly-dynamic manuveurs for various legged robots such as iCub, Talos and ANYMal. 

As a locally optimal solver, providing \emph{warm-starts} (i.e., good initial guesses) to Crocoddyl can improve its real-time performance in Model Predictive Control (MPC) setting significantly. In this work, we use the concept of memory of motion to generate the warm-starts for Crocoddyl, to avoid poor local optima while speeding up the convergence. Using HPP Loco3D~\cite{tonneau:tro:2018}, a versatile locomotion framework for legged robots, we build offline a database of humanoid walking motions. We then train function approximators using the database to generate the warm-starts for Crocoddyl.

The memory of motion concept has been used in other works, e.g. for bicopter and quadcopter~\cite{mansard2018}, serial manipulators~\cite{liu2009standing}\cite{jetchev2009trajectory}\cite{berenson2012robot}, and humanoid manipulation task~\cite{merkt2018leveraging}. However, none of them involves locomotion tasks except in~\cite{stolle2007transfer}, where a trajectory library is constructed for the LittleDog robot. Their work does not involve warm-starting an optimization solver. Instead, the sequence of joint configurations retrieved from the library is used directly, with an integral controller to correct for errors. The library is created by using a joystick to move LittleDog across the terrain. Compared to~\cite{stolle2007transfer}, our approach of using HPP Loco3D to build the library, function approximations to learn the motion, and the optimal control solver Crocoddyl to optimize the motion is more versatile and applicable to higher DoFs robots with more complex dynamics, such as the humanoid robot Talos~\cite{stasse2017talos} considered in this work.


\begin{figure}[!t]
\centering
\small
\begin{tikzpicture}[node distance = 3.5cm, auto]
    \node [block, text width=5.75em,node distance=2.75cm] (motion_generation) {Motion Generation};
    \node [block, right of=motion_generation,text width=5em,node distance=3.cm] (memory) {Memory of Motion};
    \node [block, right of=memory, text width=5.75em, node distance=3cm] (trajopt) {OC Solver};
    {\node [block, below of=motion_generation, text width=5.75em, node distance=2.cm] (trajopt2) {OC Solver};}
    {\path [line,darkgreen, dashed] (motion_generation) -- node {Hpp}(memory);}
    \path [line] (memory) -- (trajopt);
    {\path [line, blue] (motion_generation) -- node {Hpp} (trajopt2);}
    {\path [line, blue] (trajopt2) -| node[pos=0.2] {Crl} (memory);}
\end{tikzpicture}

\caption{Two approaches that are tested. In Approach 1 (dashed green), the motion generation using HPP Loco3D produces the dataset HPP, while in Approach 2 (solid blue), the dataset HPP is further optimized by the optimal control solver (Crocoddyl) to produce the dataset Crl. In each approach, the memory of motion is used for warm-starting the optimal control solver.}
\label{fig:framework}
\end{figure}
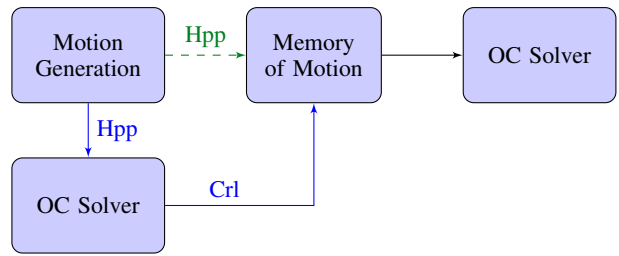

\subsection{Contribution}
Our contribution in this paper is as follows. 
Firstly, we propose a framework for learning a memory of motion to warm-start a multi-contact optimal control solver~(Fig.~\ref{fig:framework}). We generate offline a database of dynamically consistent and collision-free motions using HPP Loco3D to build the memory of motion, which is then used to warm-start the optimal control solver Crocoddyl online. We study the effect of having different solvers for the offline and online computations, and describe how we tackle this problem. 
Finally, we propose a method that learns single-step motions based on function approximations, which are then used to build multi-step motions.

The outline of the paper is as follows. In Section~\ref{sec:method}, the overall framework and the learning method are presented. Section~\ref{sec:experiments} presents the simulation results, both for the single-step and multi-step locomotions. Finally, Section~\ref{sec:conclusion} concludes the paper and discuss some ideas for the future work. 

\section{Method}
\label{sec:method}

\subsection{Problem Definition}
We consider the problem of generating whole-body locomotion of a biped robot from one location to another in a known environment and a flat terrain. The desired output consists of the robot joint configuration trajectory $\bm{q} \in \mathbb{R}^{D_q \times T}$ and the control input trajectory $\bm{u} \in \mathbb{R}^{D_u \times T}$, where $D_q, D_u, T$ are the joint configuration dimension, the control dimension, and the number of time steps. $\bm{q}$ includes both the trajectory of the root (i.e., the hip joint) and the joint positions, while $\bm{u}$ consists of the joint torques at each joint. The joint velocity and acceleration trajectories can also be included, but we only consider the joint configuration and the control trajectory here. To generate these trajectories, we use a fast optimal control solver Crocoddyl as the online solver. Our aim in this work is to provide good warm-starts to Crocoddyl using a \emph{memory of motion} such that the number of solver iterations to reach convergence can be reduced. 

\subsection{Overall Framework}

In the proposed memory of motion approach, the choice of the offline and online solver is crucial. The offline solver is to build the database of motions, while the online solver is to control the robot in real time. The online solver has to be fast and efficient, hence it is usually only locally optimal, such as Crocoddyl in this work. The offline solver, on the other hand, is used to generate the database offline, so the speed is less important. There are two alternatives: either the offline solver is the same as the online solver, or a different one. The advantage of choosing the same solver is that the motion in the database would then have the same characteristics (e.g., optimality) as the online solver, which is desirable. However, since the solver would only be locally optimal, it is difficult to use it for generating a good database without providing warm-starts to the solver.

The second alternative uses a different solver for building the database, e.g. a more global planner that can solve various tasks without requiring warm-starts. This is the approach that we take, as can be seen in Fig.~\ref{fig:framework}. HPP Loco3D~\cite{carpentier2017multi}, a locomotion framework for multi-contact locomotion, is used as the offline solver. The framework is versatile and has been applied to both biped and quadruped robots. HPP Loco3D can compute the sequence of stable contacts to achieve the locomotion task, which cannot be done yet in Crocoddyl. We use HPP Loco3D to generate motion samples corresponding to various tasks and store them as the memory of motion. 

However, the issue with this approach is that the motion in the database might not be considered \emph{optimal} by Crocoddyl, because HPP Loco3D does not properly regulate angular momentum and uses heuristics for defining the swing-foot trajectories. Indeed, there are  qualitatively different motion characteristics between HPP Loco3D and Crocoddyl. The former generates conservative and stable motions, while the latter uses the full dynamics that optimally reduces the joint torques and contact forces. Therefore, warm-starting Crocoddyl using HPP Loco3D output will require additional iterations during the online computation to refine the motion according to its optimality criteria, which is undesirable. 

We overcome this problem by leveraging both HPP Loco3D and Crocoddyl for building the memory. That is, we take the motion samples generated by HPP Loco3D and optimize them using Crocoddyl. The resulting output is then saved as the new memory of motion. By doing this, we combine the benefits of both frameworks: we can have a sequence of stable footholds (HPP Loco3D) with optimal whole-body motions (Crocoddyl). We will demonstrate that this will yield improvement in the quality of the warm-starts. In this work, we refer to the database containing the HPP Loco3D motion samples as Database $Hpp$, and the one containing the optimized Crocoddyl motion samples as Database $Crl$. 

In what follows, we describe in details HPP Loco3D and Crocoddyl frameworks.

\subsubsection{HPP Loco3D}
The planning framework proposed in HPP Loco-3D generates dynamically consistent and collision free multi-contact whole-body motion for legged robot.
It takes as input the model of the environment and an initial and goal poses of the robot's root.
Optionally, some additional constraints may be specified, such as velocity bounds or a set of initial and final contact positions. This framework decouples the motion planning problem into several sub-problems to be solved sequentially. First, the guide planning produces a rough initial guess of the root path of the robot~\cite{tonneau:tro:2018}\cite{Fernbach:iros17}.
Then, a contact planner produces a feasible sequence of contacts following the root path~\cite{tonneau:tro:2018}\cite{Fernbach:iros18}. 
After that, an optimal centroidal trajectory satisfying the centroidal dynamic constraints for the given contact points is computed~\cite{ponton2018time}. 
Finally, a second order inverse dynamic solver\footnote{https://github.com/stack-of-tasks/tsid} generates a whole-body motion, following the references of the contact locations and the centroidal trajectory. 

\subsubsection{Crocoddyl}
It is a framework for multi-contact optimal control. Given a predefined sequence of contacts, it computes efficiently the state trajectory and control policy by using sparse analytical derivatives and by exploiting the problem structure inherited from the dynamic programming principle. Its optimal control algorithm, called feasibility-prone differential dynamic programming (FDDP), has a great globalization strategy and similar numerical behavior to multiple-shooting methods~\cite{mastalli2020}. During the numerical optimization, FDDP computes the search direction and length through backward and forward passes, respectively. Unlike the classical DDP, the backward pass accepts infeasible state-control trajectories which is a critical aspect to warm-starting the solver from the memory of motion; the forward pass simulates properly infeasible search direction, obtained in the backward pass, which improves the algorithm exploration.

\subsection{Learning Strategies}
\label{sec:learning_strategies}
One important question that needs to be considered is, what do we expect the memory to learn? Should it learn directly how to move from one location to another? while this is indeed possible, such method does not generalize very well, even in a given environment. Instead, we decompose the data into single-step motions, and we retrieve a multi-step trajectory as a combination of single-step motions. 

\subsubsection{Learning Single-step Motion}
Following our previous work~\cite{lembono2019}, we formulate the problem of learning the single-step motion as a regression problem to approximate the mapping $f:\bm{x} \rightarrow \bm{y}$, where $\bm{x}$ is the task and $\bm{y}$ is the corresponding trajectory output. We separate the database into left-leg and right-leg movements, as this yields better results than combining both and let the memory learns how to choose the leg. Each motion starts with the root position at the origin.\footnote{Without any loss of generality, as we can always transform the coordinate system} The task is defined as the initial and goal foot poses, $\bm{x} = [\bm{c}_{l0}, \bm{c}_{r0}, \bm{c}_{*T}] \in \mathbb{R}^9$, where $\bm{c} \in \mathbb{R}^3$ is the foot pose (2D position and orientation), the subscripts $l,r$ correspond to the left and right foot, while $*$ is either $l$ for the left-leg or $r$ for the right-leg database. $T$ is the final time step. $\bm{y}$ can be either the joint configuration trajectory $\bm{q} \in \mathbb{R}^{D_q \times T}$ or the control trajectory $\bm{u} \in \mathbb{R}^{D_u \times T}$. We then apply dimensionality reduction and function approximation techniques to solve the regression problem.

Since the dimension of $\bm{y} \in \mathbb{R}^{D \times T}$ is high, we need to find a smaller representation of $\bm{y}$. First, we represent the trajectory of each dimension of $\bm{y}$ as the weight of radial basis functions (RBFs), as commonly done in probabilistic movement primitives~\cite{paraschos2013probabilistic}\cite{Calinon19MM}. 
Let $\bm{y}_i \in \mathbb{R}^T$ be the trajectory of the $i_{\text{th}}$ dimension of $\bm{y}$. We can define the basis matrix $\bm{\Phi}$ as $[\bm{\phi}_0, \dots, \bm{\phi}_{K-1}] \in \mathbb{R}^{T \times K}$, where $\bm{\phi}_k \in \mathbb{R}^T$ is the $k_{th}$ basis function, defined as a Gaussian function centered at the $k_{\text{th}}$ mean. The means are distributed equally along the time axis $T$, whereas the variance is chosen to be equal for all the basis functions and to have sufficient overlap. 
$\bm{y}_i$ can then be represented as the weights of these basis functions, 
\begin{equation}
\bm{y}_i = \bm{\Phi} \bm{w}_i,
\end{equation}
where $\bm{w}_i \in \mathbb{R}^K$ can be computed by standard linear least squares algorithm. This reduces the number of variables for each dimension from $T$, which is usually large, to $K$ which can be much smaller. We can then stack the weights corresponding to all dimensions of $\bm{y}$ and obtain $\bm{w} = [\bm{w}_0, \dots, \bm{w}_i, \dots, \bm{w}_{D-1}] \in \mathbb{R}^{DK}$. Each $\bm{y}$ has the corresponding weight vector $\bm{w}$. 


Now let's consider all the $N$ samples in the database. We can apply principal component analysis (PCA) to further reduce the dimension of $\bm{w}$ over this database by keeping only $M$ principal components to obtain $\hat{\bm{y}} \in \mathbb{R}^{M}$. We have thus reduced the dimension of $\bm{y}$ from $DT$ to $M$. Inverse transformation from $\hat{\bm{y}}$ to $\bm{y}$ is a matrix multiplication that can be performed quickly. The regression problem then becomes $f:\bm{x} \rightarrow \hat{\bm{y}}$. To solve the regression problem, we consider two function approximation techniques: Gaussian process regression (GPR) and Gaussian mixture regression (GMR). 

\subsubsubsection{Gaussian Process Regression (GPR)}

GPR~\cite{rasmusse:book:2006} is a non-parametric method which improves its accuracy as the number of datapoints increases. Given the database $\{\bm{X}, \bm{Y}\}$, GPR assigns a Gaussian prior to the joint probability of $\bm{Y}$, i.e., $p(\bm{Y}|\bm{X})= \mathcal{N}\big(\bm{\mu}(\bm{X}), \bm{K}(\bm{X},\bm{X})\big)$. $\bm{\mu}(\bm{X})$ is the mean function and $\bm{K}(\bm{X},\bm{X})$ is the covariance matrix constructed with elements $\bm{K}_{ij} =  {k}(\bm{x}_i,\bm{x}_j)$, where ${k}(\bm{x}_i,\bm{x}_j)$ is the kernel function that measures the similarity between the inputs $\bm{x}_i$ and $\bm{x}_j$. In this paper we use RBF as the kernel function, and the mean function $\bm{\mu}(\bm{X})$ is set to zero, as usually done in GPR. 

To predict the output $\bm{y}^*$ given a new input $\bm{x}^*$, GPR constructs the joint probability distribution of the training data and the prediction, and then conditions on the training data to obtain the predictive distribution of the output, $p({\bm{y}}^* |~\bm{x}^*) \sim \mathcal{N}(\bm{m},\bm{\Sigma})$, where $\bm{m}$ is the posterior mean computed as
\begin{equation}
\bm{m} = \bm{K}({\bm{x}}^*,\bm{X})\, \bm{K}^{-1}(\bm{X},\bm{X})\,  \bm{Y}(\bm{X}), \label{eq:post_mean}
\end{equation}
and $\bm{\Sigma}$ is the posterior covariance which provides a measure of uncertainty on the output. In this work we simply use the posterior mean $\bm{m}$ as the output, i.e., $\bm{y}^* = \bm{m}$.

\subsubsubsection{Gaussian Mixture Regression (GMR)}
Given the training database $\{\bm{X},\bm{Y}\}$ GMR constructs the joint probability of $(\bm{x},\bm{y})$ as a mixture of Gaussians,
\begin{equation}
p(\bm{x},\bm{y}) = \sum_{k=1}^K \pi_k \, \mathcal{N}(\bm{\mu}_k,\bm{\Sigma}_k),
\end{equation}	
where $\pi_k$, $\bm{\mu}_k$, and $\bm{\Sigma}_k$ are the $k$-th component's mixing coefficient, mean, and covariance, respectively. Let $\bm{\theta} = \{\pi_k,\bm{\mu}_k, \bm{\Sigma}_k\}_{k=1}^K$, denoting the GMR parameters to be determined from the data. 

We can decompose $\bm{\mu}_k$ and $\bm{\Sigma}_k$ according to $\bm{x}$ and $\bm{y}$ as
\begin{equation}
\bm{\mu}_k = \begin{pmatrix} \bm{\mu}_{k,x} \\ \bm{\mu}_{k,y} \end{pmatrix} \quad \text{and} \quad \bm{\Sigma}_k = 
\begin{pmatrix}
\bm{\Sigma}_{k,xx} & \bm{\Sigma}_{k,xy} \\
\bm{\Sigma}_{k,yx} & \bm{\Sigma}_{k,yy}
\end{pmatrix}.
\end{equation}
Given a query $\bm{x}^*$, the predictive distribution of $\bm{y}$ can then computed by conditioning on $\bm{x}^*$, 
\begin{equation}
p(\bm{y}|~\bm{x}^*,\bm{\theta}) = \sum_{k=1}^K p(k|~\bm{x}^*,\bm{\theta}) \, p(\bm{y}|~k,\bm{x}^*,\bm{\theta}), 
\label{eq:gmr_pred_sup}
\end{equation}
where $p(k|~\bm{x}^*,\bm{\theta})$ is the probability of $\bm{x}^*$ belonging to the $k$-th component, 
\begin{equation}
p(k|~\bm{x}^*,\bm{\theta}) = \frac{\pi_k \, \mathcal{N}(\bm{x}^* |~ \bm{\mu}_{k,x}, \bm{\Sigma}_{k,xx})}{\sum_{i=1}^{K} \pi_i \, \mathcal{N}(\bm{x}^* |~ \bm{\mu}_{i,x}, \bm{\Sigma}_{i,xx}) },
\label{eq:prob_k}
\end{equation}
and $p(\bm{y}|~k,\bm{x}^*,\bm{\theta}$) is the predictive distribution of $\bm{y}$ according to the $k$-th component, 
\begin{multline}
p(\bm{y}|~k,\bm{x}^*,\bm{\theta}) = \mathcal{N}\Big(\bm{\mu}_{k,y} + \bm{\Sigma}_{k,yx} {\bm{\Sigma}^{-1}_{k,xx}}(\bm{x}^* - \bm{\mu}_{k,x}), \\
 \bm{\Sigma}_{k,yy} - \bm{\Sigma}_{k,yx} {\bm{\Sigma}^{-1}_{k,xx}} \bm{\Sigma}_{k,xy}\Big),
\label{eq:pred_k}
\end{multline}
which is a	 Gaussian distribution with the mean being linear in $\bm{x}^*$. The resulting predictive distribution \eqref{eq:gmr_pred_sup} is then a mixture of Gaussians. The point prediction $\bm{y}^*$ can be obtained from this distribution by applying moment matching to the distribution in \eqref{eq:gmr_pred_sup} to approximate it by a single Gaussian, and use the mean of the Gaussian as the desired output $\bm{y}^*$, see~\cite{Calinon19MM,Ghahramani94} for details.

\subsubsection{Constructing Multi-step Motion}
\label{sec:multistep}
To use the single-step motion for generating multi-step motions, we have to transform the coordinate system at each step. Let's first assume that we have the planned sequence of contacts (i.e., foot poses), $\{\bm{C}_i\}_{i=0}^{P-1}$, where $\bm{C}_i = (\bm{c}_{li}, \bm{c}_{ri})$ is the contacts at $i_{\text{th}}$ footstep and $P$ is the total number of footsteps. Assume, without loss of generality, that the motion starts at zero root position horizontally. The first step can be obtained by querying the single-step memory directly to move from $\bm{C}_0$ to $\bm{C}_1$, obtaining $\bm{y}^0$. To move from $\bm{C}_1$ to $\bm{C}_2$, we first need to shift the coordinate system such that the current root (based on the last configuration at $\bm{y}^0$) is in zero horizontal position. The motion from $\bm{C}_1$ to $\bm{C}_2$ can then be queried from the memory to obtain $\bm{y}^1$, and this is iterated until $\bm{C}_{P-1}$ to obtain $\{\bm{y}^i\}_{i=0}^{P-1}$. Finally, each trajectory $\bm{y}^i$ is transformed back to the original coordinate system and concatenated as a single trajectory $\bm{y}$. 

The contact sequence $\{\bm{C}_i\}_{i=0}^{P-1}$ can be obtained from another planner, such as RB-PRM~\cite{tonneau:tro:2018}. The alternative is to also learn it from the database. In this work we assume that the contact sequence is already given, and the contact sequence learning will be considered in future work.

\section{Experiments}
\label{sec:experiments}

To evaluate the proposed approach, we conduct several experiments with the humanoid robot Talos in simulation. The robot joint configuration consists of 3 DoFs root position, 3 DoFs root orientation (described in quaternion), and 32 joint angles ($D_q = 39$), while the control input consists of 32 joint torques ($D_u=32$). First, HPP Loco3D is used to generate $N=1200$ samples of one-step motion (right-leg and left-leg movement in equal proportions), starting from the initial contact $(\bm{c}_{l0}, \bm{c}_{r0})$ to the goal contact $(\bm{c}_{lT}, \bm{c}_{rT})$. One sample thus consists of $\{(\bm{c}_{l0}, \bm{c}_{r0}), (\bm{c}_{lT}, \bm{c}_{rT}), \bm{q}, \bm{u}\}$. These are stored as Database Hpp.\footnote{The database is available at \url{https://github.com/MeMory-of-MOtion/docker-loco3d.}} Next, each sample is optimized using Crocoddyl, and the resulting samples are stored as Database Crl. The cost function in Crocoddyl consists of state and control regularization (around a standing pose and zero, respectively), and contact placement. Since we need high-quality database for the memory of motion, we use a small convergence threshold of $10^{-5}$ and the maximum number of iterations is set to be $50$. The time interval in HPP Loco3D is 1~ms and 40~ms, respectively, so the HPP Loco3D data is subsampled to Crocoddyl's interval for the optimization. Both databases will be used and the performance will be compared. The databases are split into the training and the test dataset, with $N_{\text{train}} =1000$ and $N_{\text{test}} = 200$. We applied RBF and PCA to reduce the dimensions of $\bm{q}$ and $\bm{u}$ with $K=60$ and $M=60$, as determined empirically. 


We proceed as follows. First, we evaluate the accuracy performance of GPR and GMR in approximating the mapping $f$. The warm-starts generated by GPR and GPR are then compared to the cold-start in terms of the Crocoddyl performance, i.e., the number of iterations until convergence and the resulting trajectory cost. Finally, we also compare the result of warm-starts using only $\bm{q}$ to using both $\bm{q}$ and $\bm{u}$. For all comparisons, we use both the databases Hpp and Crl and compare their performance.

\subsection{Comparing GPR and GMR Accuracy}

\begin{figure*}[t!]
\captionsetup[subfigure]{labelformat=empty}
\centering
\subfloat[GPR][]
{
\fbox{\includegraphics[width=0.25\columnwidth]{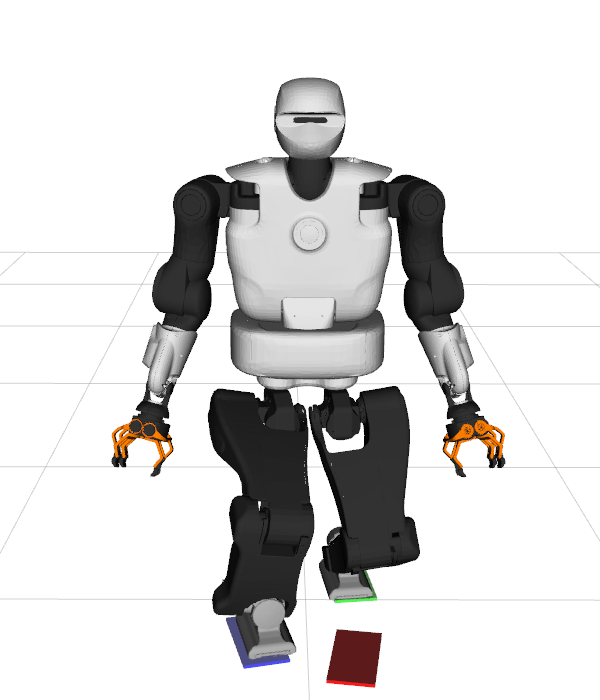}}
\fbox{\includegraphics[width=0.25\columnwidth]{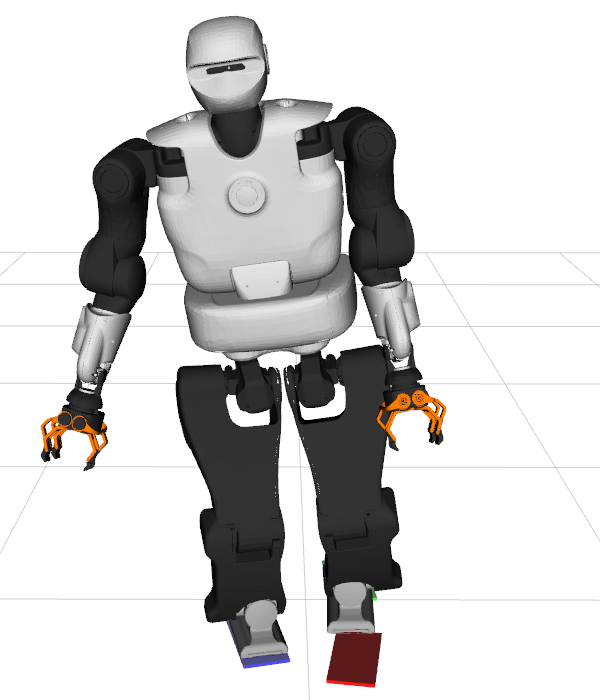}}
\fbox{\includegraphics[width=0.25\columnwidth]{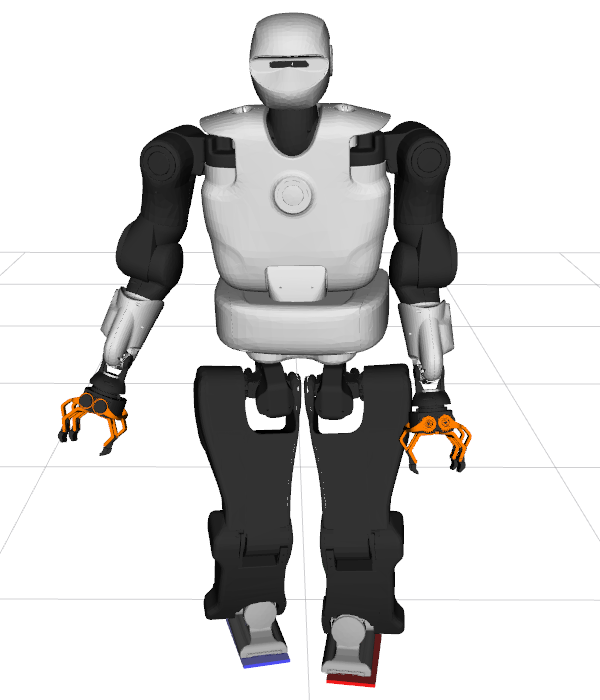}} \quad \quad
\fbox{\includegraphics[width=0.25\columnwidth]{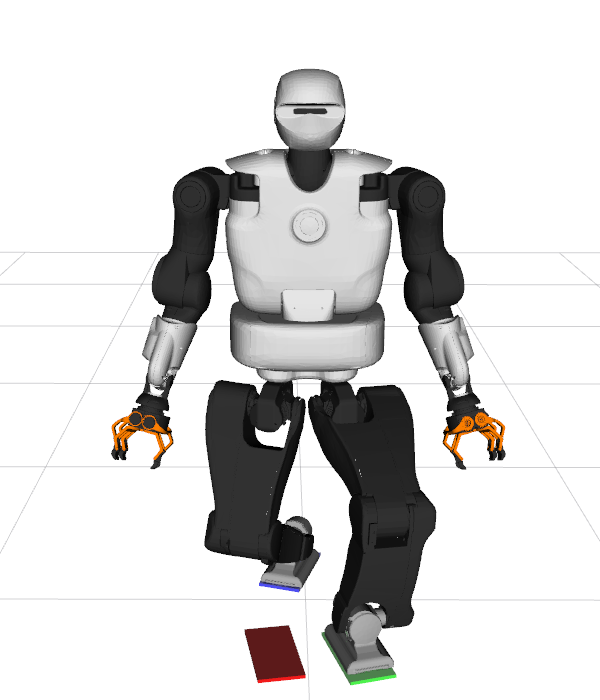}}
\fbox{\includegraphics[width=0.25\columnwidth]{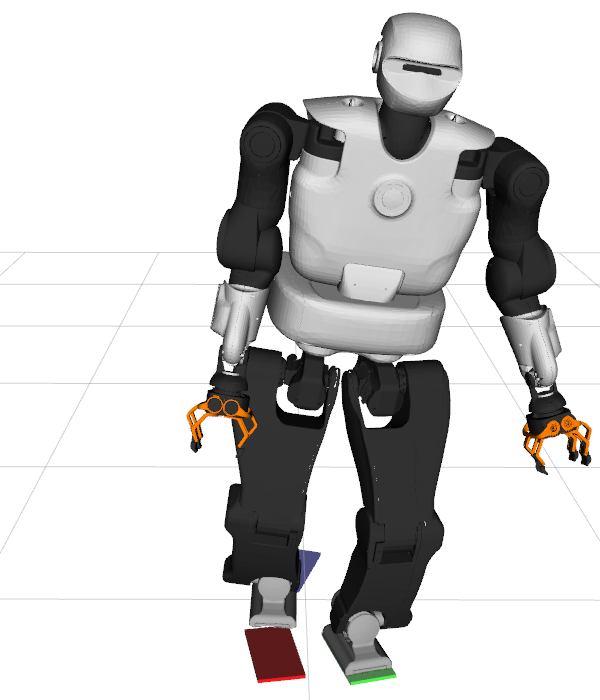}} 
\fbox{\includegraphics[width=0.25\columnwidth]{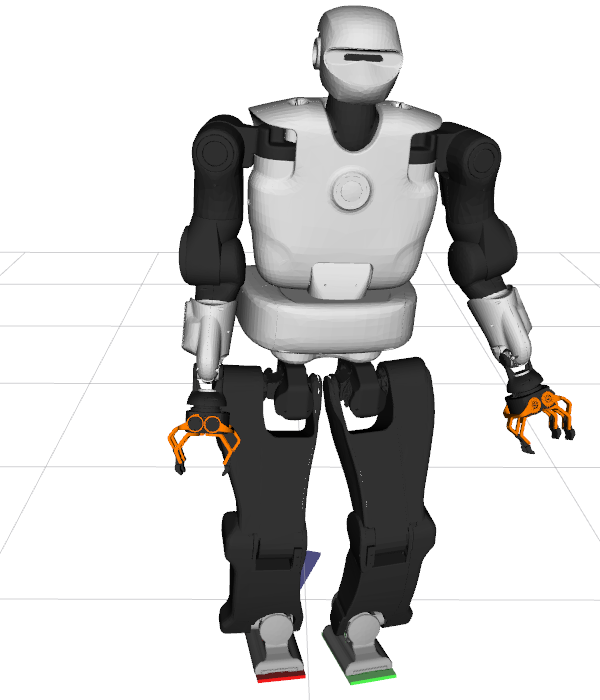}}
\label{fig:gpr_pred}
}\\[-2ex]
\subfloat[][]
{
\fbox{\includegraphics[width=0.25\columnwidth]{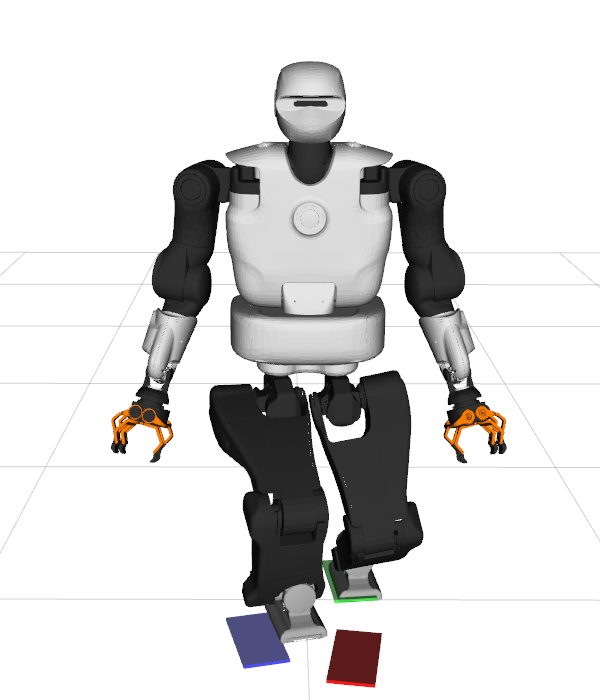}}
\fbox{\includegraphics[width=0.25\columnwidth]{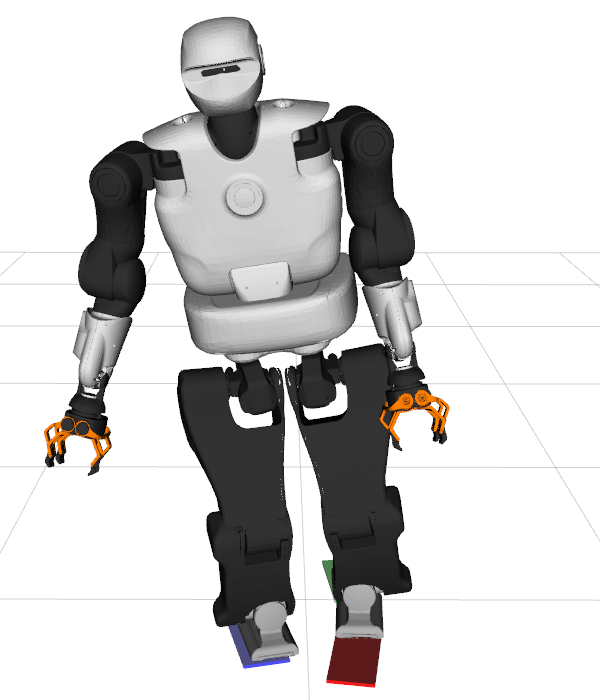}}
\fbox{\includegraphics[width=0.25\columnwidth]{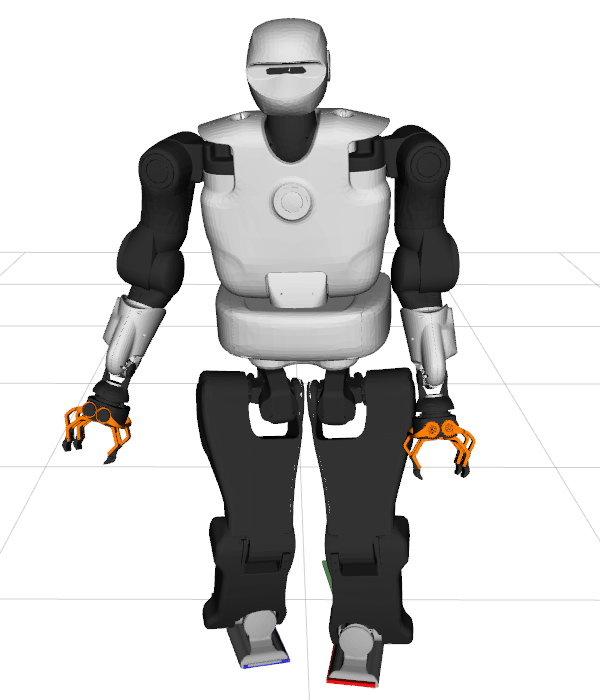}}\quad \quad
\fbox{\includegraphics[width=0.25\columnwidth]{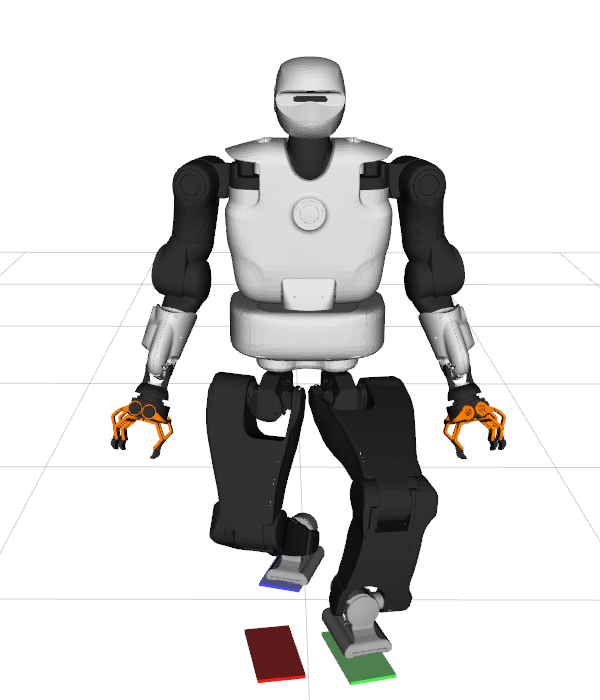}}
\fbox{\includegraphics[width=0.25\columnwidth]{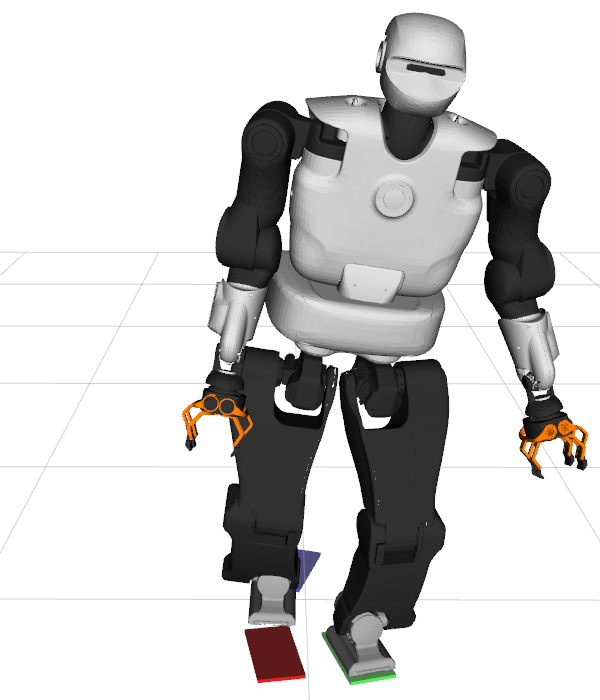}} 
\fbox{\includegraphics[width=0.25\columnwidth]{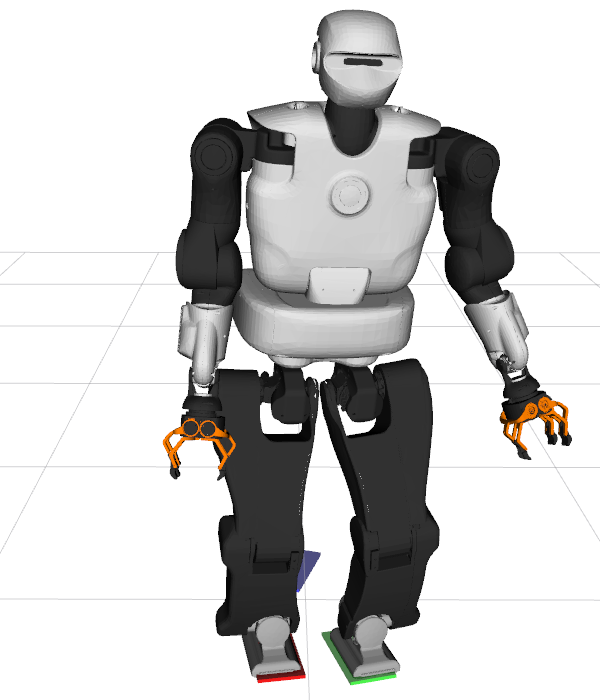}}
\label{fig:gmr_pred}
}\\[-2ex]
\subfloat[][]
{
\fbox{\includegraphics[width=0.25\columnwidth]{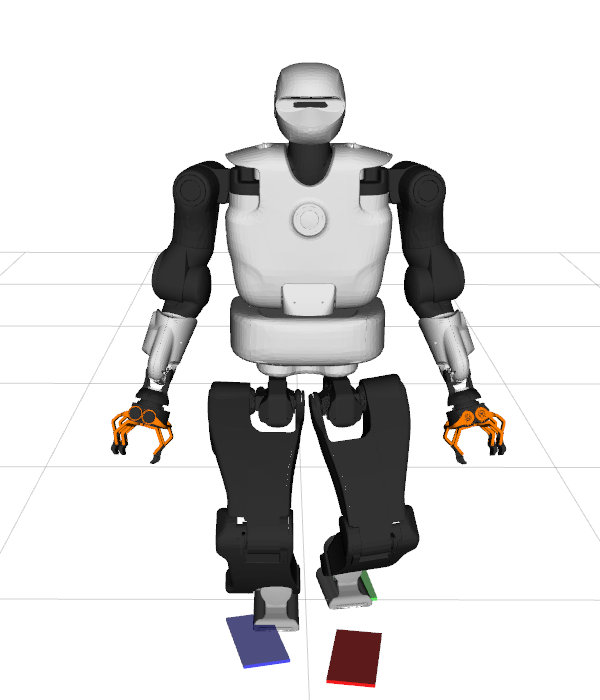}}
\fbox{\includegraphics[width=0.25\columnwidth]{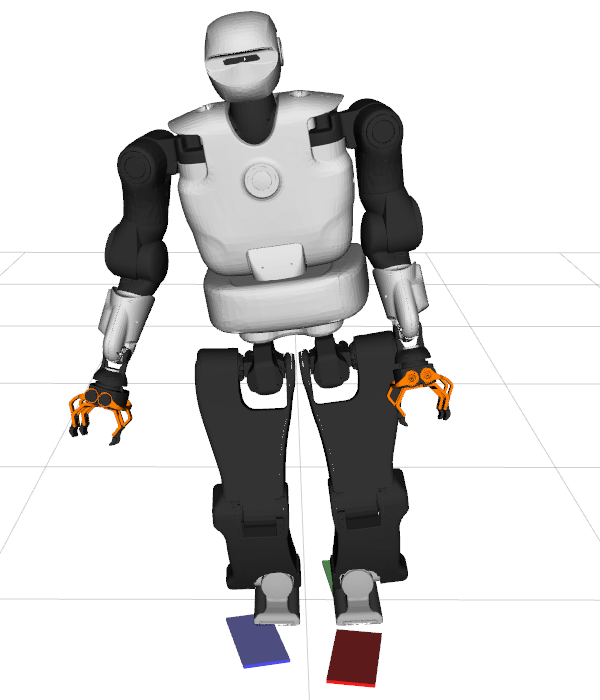}}
\fbox{\includegraphics[width=0.25\columnwidth]{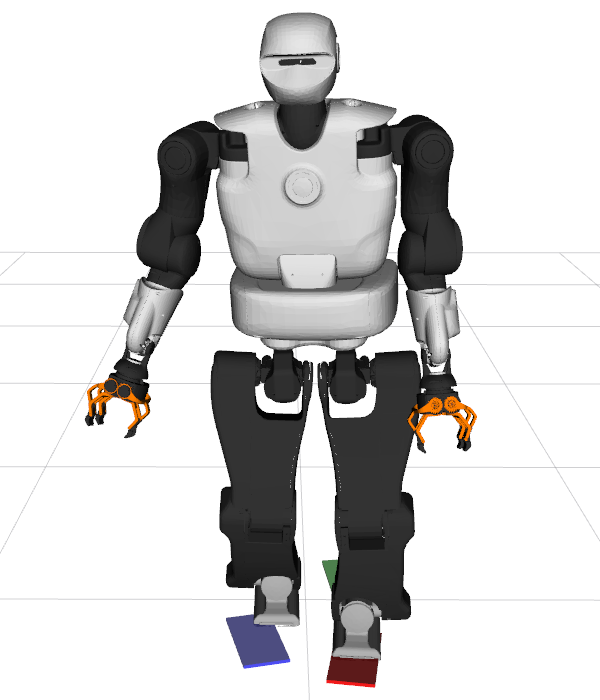}}\quad \quad
\fbox{\includegraphics[width=0.25\columnwidth]{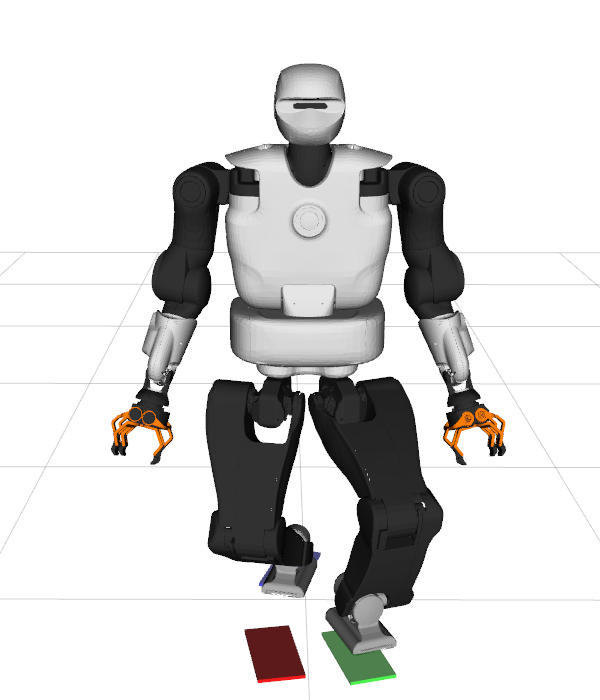}}
\fbox{\includegraphics[width=0.25\columnwidth]{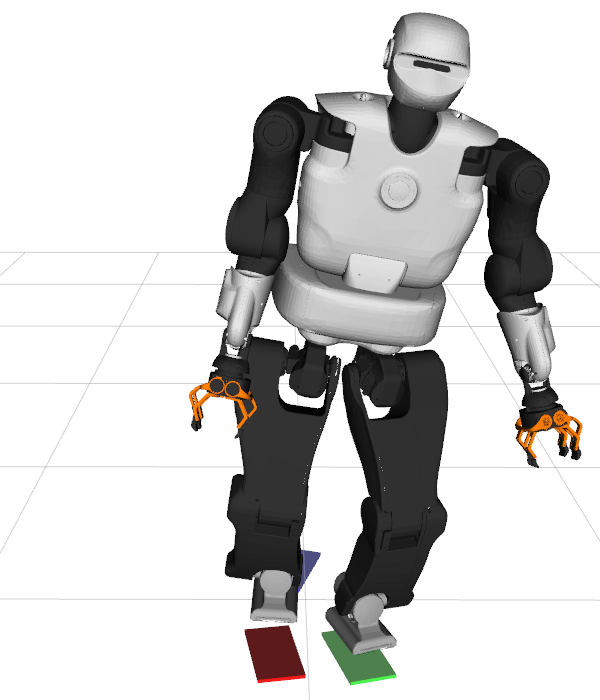}} 
\fbox{\includegraphics[width=0.25\columnwidth]{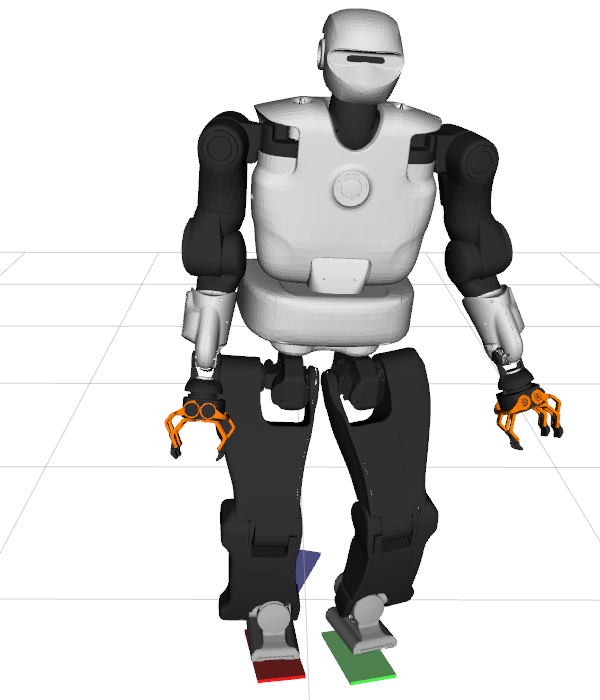}}
\label{fig:knn_pred}
}
\caption{Examples of predicting single-step motions. Warm-start produced by \emph{Top:}~GPR, \emph{Middle:}~GMR, and \emph{Bottom:}~$k$-NN. \emph{Left:} left foot movement. \emph{Right:} right foot movement. The green, blue and red box correspond to the initial left, the initial right, and the goal contact pose, respectively. }
\label{fig:compare_pred}
\end{figure*}		

We train GPR and GMR on the reduced-dimension dataset $\{\bm{x},\hat{\bm{y}}\}$ for both databases Hpp and Crl with $N_{\text{train}}$ samples. The task $\bm{x}$ is defined as the initial and goal foot poses, $\bm{x} = [\bm{c}_{l0}, \bm{c}_{r0}, \bm{c}_{*T}]$ where $*=l$ for the left-leg movement and $*=r$ for the right-leg movement. The output $\bm{y}$ is defined here as the joint configuration trajectory $\bm{q}$, and $\hat{\bm{y}}$ is its smaller dimension representation. For each $\bm{x}$ in the $N_{\text{test}}$ samples, we use GPR and GMR to predict the corresponding trajectory $\bm{y}$, and the accuracy is evaluted against the true trajectory in the database. The trajectory error~($rad$) is calculated as the difference between the true and predicted trajectory, i.e. $\frac{1}{N_{\text{test}}}\sum_{i=1}^{N_{\text{test}}} {||\bm{y}^i - \tilde{\bm{y}}^i||}_2$, whereas the contact error~($m$) is defined as the difference between the foot poses of the true and predicted trajectory, $\frac{1}{N_{\text{test}}}\sum_{i=1}^{N_{\text{test}}} ||\bm{C}^i - \tilde{\bm{C}}^i||_2$. 

The results can be seen in Table~\ref{tab:func_accuracy} (averaged over the left and right leg movements). We compare against $k$-NN with $k=1$ as a baseline, to demonstrate that the proposed algorithm indeed generalizes well and the good performance is not due to having a very dense database. GPR overall has the lowest errors in both criteria and both databases. GMR has higher errors than GPR but still outperforms the baseline $k$-NN by a large margin. Some example of the motions can be found in Fig. \ref{fig:compare_pred}. We can see that motion predicted by GPR fit the desired contact locations very well.

Furthermore, for the subsequent results we will use the subscripts Hpp and Crl for the function approximators trained on the databases Hpp and Crl, respectively. 


\renewcommand{\arraystretch}{1.}
\begin{table}[!t]
	\centering
	\addtolength{\tabcolsep}{-1.5pt}
      \caption{Comparing The Accuracy of GPR and GMR}
      
	\begin{tabular}{ l  c  c  c  c}
		\toprule
		 & \multicolumn{2}{c}{\textbf{Database Hpp}} & \multicolumn{2}{c}{\textbf{Database Crl}} \\
		\textbf{Method} & \textbf{Traj. Err.} & \textbf{Contact Err.} & \textbf{Traj. Err.} & \textbf{Contact Err.}  \\  
       \midrule
		\rowcolor{LightCyan} GPR & 9.53 $\pm$ 4.63 &	 0.07 $\pm$ 0.03 & 	 13.04 $\pm$ 6.15 	 & 0.04 $\pm$ 0.02	\\
		GMR &  12.39 $\pm$ 5.00 	& 0.13 $\pm$ 0.05  & 	 18.51 $\pm$ 6.80 	 & 0.09 $\pm$ 0.04	\\
		$k$-NN &  18.78 $\pm$ 4.96 	& 0.49 $\pm$ 0.15	 & 	 29.93 $\pm$ 9.02 	 & 0.51 $\pm$ 0.10 \\
		\bottomrule
	\end{tabular}
	\label{tab:func_accuracy}
\end{table}

\subsection{Single-step Motion: Warm-start vs Cold-start}
\label{sec:onestep_result}

In this section we compare the results of warm-starting Crocoddyl using the function approximators against the cold-start, i.e., using zero initial guess. For each $\bm{x}$ in the $N_{\text{test}}$ samples, GPR and GMR are queried to obtain the initial guess $\bm{y}$, defined here as the joint configuration trajectory $\bm{q}$. We do not predict the trajectory of the control input here, but instead calculate it from $\bm{q}$ assuming quasi-static movement. This computed control input trajectory is denoted as $\bm{u}_0$. The initial guess is used to warm-start Crocoddyl, and the result is compared against the cold-start. We also compare the effect of using Database Hpp and Crl. We use a large convergence threshold ($10^{-2}$) for Crocoddyl here, based on the assumption that at online computation a very refined optimal motion is not really necessary. The number of iterations is also limited to 20. The query time is $\sim$5ms for GPR and $\sim$10ms for GMR in python implementation. 

The results can be seen in Table~\ref{tab:onestep_warmcold}. The solver is considered successful if it finds a feasible trajectory within the maximum number of iterations. We see from the table that the warm-starts results consistently outperform the cold-starts; it justifies our assumption that warm-starting Crocoddyl can speed-up the computation time for online MPC. 

Although in Table~\ref{tab:func_accuracy} we see that the accuracy of GPR outperforms GMR quite substantially, the warm-starting performance in Table~\ref{tab:onestep_warmcold} turns out to be very similar, with GPR having very slightly better performance. This is due to the fact that the initial guesses produced by GPR and GMR still go through an optimization process in Crocoddyl, and hence the small difference between the predictions does not affect the results much. This means that while in standard regression tasks the accuracy is highly important, it may be less important in tasks such as producing warm-starts, as long as the predicted initial guesses are sufficiently close to the optimal solutions. 

Finally, we compared the results between the function approximators trained on Database Hpp (GPR$_{\text{Hpp}}$, GMR$_{\text{Hpp}}$) and Crl (GPR$_{\text{Crl}}$, GMR$_{\text{Crl}}$). Those trained on the database Crl have lower number of iterations, which justify our step of optimizing the HPP Loco3D dataset by Crocoddyl. The dataset in Crl contain motion samples that are optimal according to Crocoddyl, and therefore they perform better in warm-starting Crocoddyl.


\renewcommand{\arraystretch}{.9}
\begin{table}[!t]
	\centering
      \caption{Comparing Warm-start vs Cold-start: Single-step}
      
	\begin{tabular}{ l  c c  c }
		\toprule
		\textbf{Method} & \textbf{Success Rate} & \textbf{Cost} & \textbf{Num. Iteration} \\  
       \midrule
		Cold-start     	 & 98.50 & 	 55.56 $\pm$8.32 & 	 9.52 $\pm$ 3.94\\
		GPR$_{\text{Hpp}}$ & 99.00 & 	 55.27 $\pm$8.01 & 	 5.01 $\pm$ 0.74\\
		GMR$_{\text{Hpp}}$    & 100.00 & 	 55.31 $\pm$7.99 & 	 4.85 $\pm$ 0.66\\
		\rowcolor{LightCyan} GPR$_{\text{Crl}}$ & 100.00 & 	 55.32 $\pm$7.99 & 	 3.04 $\pm$ 0.41\\
		GMR$_{\text{Crl}}$   	 & 100.00 & 	 55.32 $\pm$7.99 & 	 3.06 $\pm$ 0.45	\\
			\bottomrule
	\end{tabular}
	\label{tab:onestep_warmcold}
\end{table}

\begin{figure*}[t!]
\centering
\subfloat[GPR][]
{
\fbox{\includegraphics[width=0.35\columnwidth]{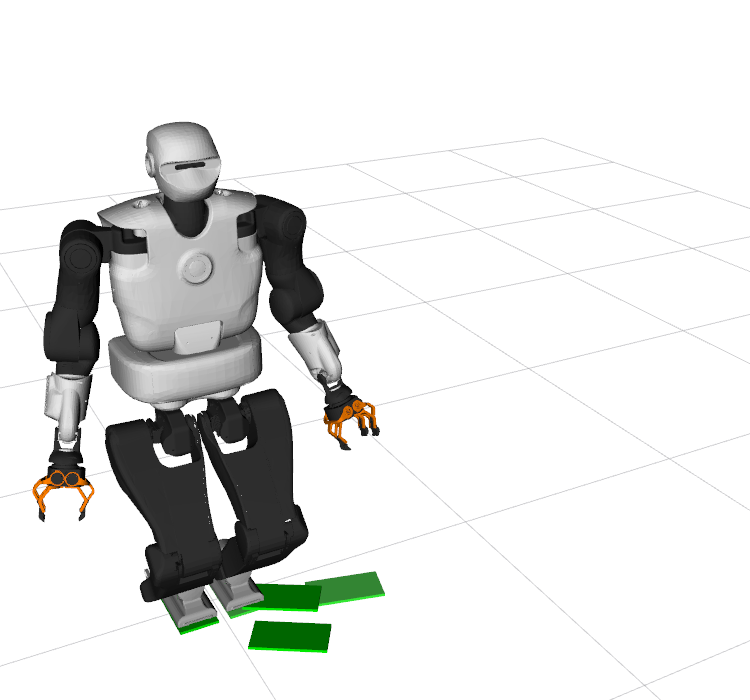}}
\fbox{\includegraphics[width=0.35\columnwidth]{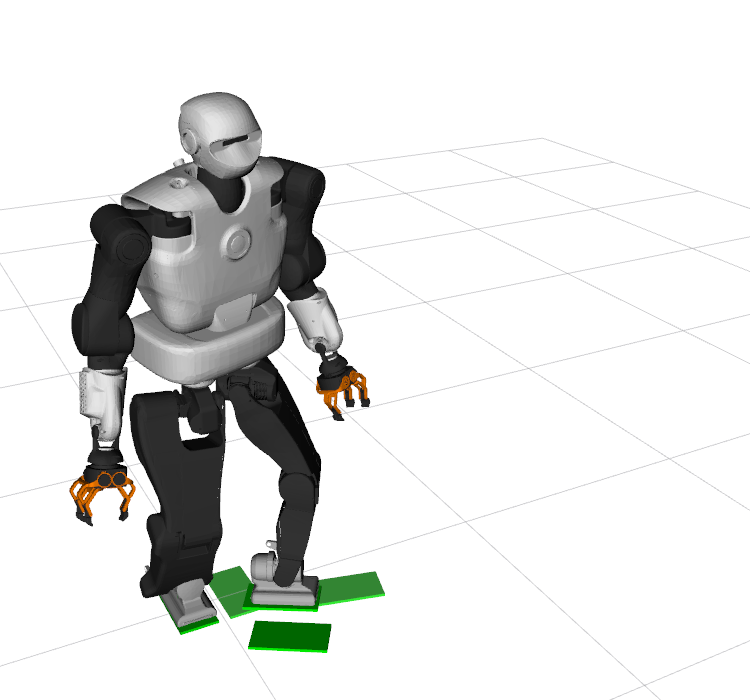}}
\fbox{\includegraphics[width=0.35\columnwidth]{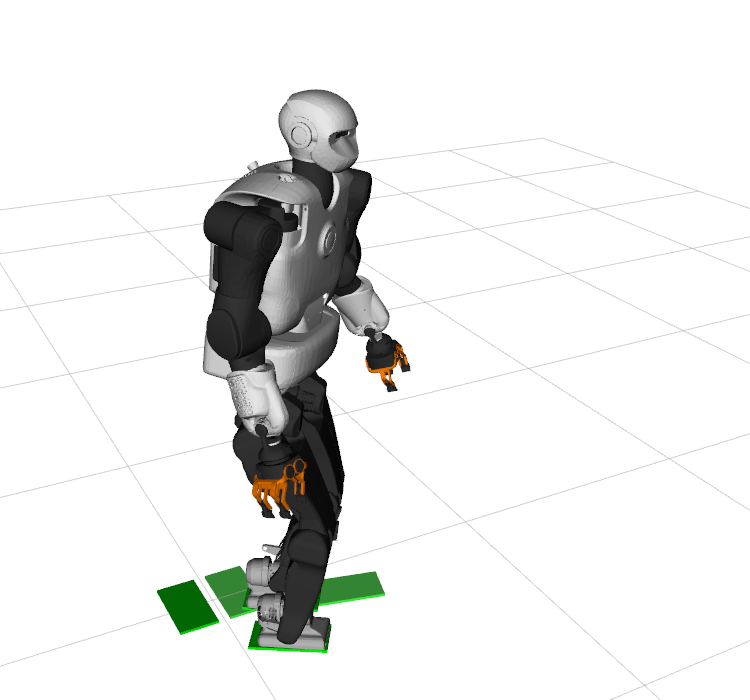}}
\fbox{\includegraphics[width=0.35\columnwidth]{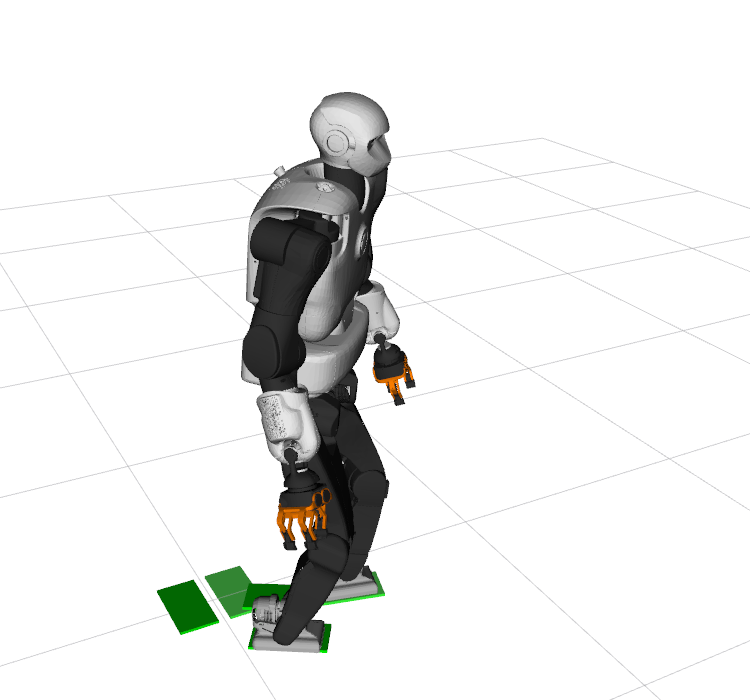}}
\label{fig:gpr_pred}
}\\
\caption{Examples of predicting a multi-step motion by GPR.}
\label{fig:multistep}
\end{figure*}

\subsection{Single-step Motion: Evaluating Warm-starts Components}

In Section \ref{sec:onestep_result}, we only predict the joint configuration trajectory $\bm{q}$ for warm-starting Crocoddyl while the control trajectory $\bm{u}$ is computed based on the predicted $\bm{q}$. In this section we evaluate the different performances if we also predict $\bm{u}$, or use zero control trajectory as warm-start. We train two different GPRs for the prediction, one for predicting $\bm{q}$ and the other one for predicting $\bm{u}$. 

Table \ref{tab:onestep_comps} shows the comparison results. $(\bm{q},\bm{u}_0)$ denotes the warm-starts using the predicted $\bm{q}$ while the control trajectory is computed as $\bm{u}_0$, the same as in the previous subsection. $(\bm{q})$ denotes the warm-starts using only the joint configuration trajectory while the control trajectory is set to be zero. While the cost remains the same, the number of iterations increases when the control trajectory is omitted from the warm-starts. $(\bm{q},\bm{u})$ denotes the warm-starts using both predicted joint configuration and control trajectory, which has similar results to  $(\bm{q},\bm{u}_0)$ except for the slightly lower number of iterations. Finally, $(\bm{u})$ denotes the warm-starts using only the control trajectory while the joint configuration trajectory is set to zero, and cold-start means warm-starting using zero joint configuration and control trajectory. 

Comparing $(\bm{q},\bm{u})$ and  $(\bm{q},\bm{u}_0)$, we conclude that predicting the control trajectory from the memory does not give significant benefit as compared to computing it based on $\bm{q}$. However, control trajectory is still important to be included in the warm-starts, as omitting them in $(\bm{q})$ increases the number of iterations. Finally, comparing $(\bm{q})$ and $(\bm{u})$, we conclude that warm-starting using only the joint configuration trajectory performs better than using only control trajectory, which has higher cost and number of iterations.

\renewcommand{\arraystretch}{.9}
\begin{table}[!t]
	\centering
      \caption{Comparing Warm-start Components: Single-step}
      
	\begin{tabular}{ l  c  c c c}
		\toprule
		\textbf{Method} & \textbf{Success Rate} & \textbf{Cost} & \textbf{Num. Iteration} \\ 
       \midrule
		$(\bm{q})$   & 96.50 & 	 55.06 $\pm$8.19 & 	 5.30 $\pm$ 2.19 \\
		$(\bm{q},\bm{u}_0)$   	& 100.00 & 	 55.32 $\pm$7.99 & 	 3.04 $\pm$ 0.41\\
		\rowcolor{LightCyan} $(\bm{q},\bm{u})$  & 100.00 & 	 55.32 $\pm$7.99 & 	 2.93 $\pm$ 0.45\\
		$(\bm{u})$   	 & 97.50 & 	 55.36 $\pm$7.95 & 	 6.83 $\pm$ 2.46\\
		Cold-start   & 98.50 & 	 55.56 $\pm$8.32 & 	 9.52 $\pm$ 3.94 \\
		\bottomrule
	\end{tabular}
	\label{tab:onestep_comps}
\end{table}

\subsection{Multi-step Motion: Warm-start vs Cold-start}

Finally, we use the single-step motions as a building block for multi-step locomotion, as described in Section~\ref{sec:learning_strategies}. We use HPP Loco3D to generate $50$ contact sequences, each consists of $P=3$ footsteps. GPR$_{\text{Hpp}}$, GMR$_{\text{Hpp}}$, GPR$_{\text{Crl}}$ and GMR$_{\text{Crl}}$ generate the initial guesses of the joint configuration trajectory $\bm{q}$ while the control trajectory is computed based on $\bm{q}$, as in Section \ref{sec:onestep_result}. The warm-starts are then given to Crocoddyl, and the results are presented in Table~\ref{tab:multistep_warmcold}. Interestingly the Hpp database does not perform better than the cold-start (indeed, it is the opposite). This could be due to the fact that we build the motion using a concatenation of three single-step motions, each step starts and ends with zero velocity. The nonoptimality of the Hpp database, added with the nonoptimality of the multi-step motion strategy, seem to render the warm-start to be not useful at all. On the contrary, warm-starting using the Crl database still speeds-up the convergence, although not as much as in the single-step case (Section \ref{sec:onestep_result}). An example of multi-step locomotion warm-start produced by GPR is given in Fig. \ref{fig:multistep}. 

\renewcommand{\arraystretch}{1.}
\begin{table}[!t]
	\centering
      \caption{Comparing Warm-start vs Cold-start: Multi-Step}
      
	\begin{tabular}{ l  c c  c }
		\toprule
		\textbf{Method} & \textbf{Success Rate} & \textbf{Cost} & \textbf{Num. Iteration} \\  
       \midrule
		Cold-start     	  & 100.00 & 	 86.36 $\pm$23.16 & 	 6.17 $\pm$ 1.52\\
		GPR$_{\text{Hpp}}$  & 100.00 & 	 86.39 $\pm$23.07 & 	 6.40 $\pm$ 0.73 \\
		GMR$_{\text{Hpp}}$   	   & 100.00 & 	 86.38 $\pm$23.12 & 	 7.29 $\pm$ 1.32\\
		\rowcolor{LightCyan} GPR$_{\text{Crl}}$  & 100.00 & 	 86.47 $\pm$23.16 & 	 4.54 $\pm$ 0.55 \\
		GMR$_{\text{Crl}}$   & 100.00 & 	 86.51 $\pm$23.22 & 	 4.71 $\pm$ 0.70	\\
			\bottomrule
	\end{tabular}
	\label{tab:multistep_warmcold}
\end{table}

\section{Conclusion and Future Work}
\label{sec:conclusion}

We have presented a framework for learning a memory of motion to warm-start an optimal control solver. The proposed approach manages to reduce the average number of solver iterations from $\sim$9.5 to only $\sim$3.0 iterations for the single-step motion and from $\sim$6.2 to $\sim$4.5 iterations for the multi-step motion, while maintaining the solution's quality. 

This paper shows a preliminary result of warm-starting an optimal control solver. In the current formulation, Crocoddyl does not include constraints such as torque limit or obstacle avoidance, and we are working towards this direction. When the optimal control formulation becomes more realistic and complex, the solver would need even higher computational time, and the proposed warm-starting approach would potentially be even more useful. The final goal is to use the whole framework to control the real robot using MPC.

In this work, we decompose the multistep locomotion into single-step motions, with the initial and goal contact locations as the task that needs to be provided by some other methods. Another potential strategy is to define the task to be the final root pose instead of the contact location. The memory will then determine the contact location to reach the desired root pose. This may allow more flexibility in generating the multi-step motions, as only the root trajectory needs to be provided instead of the contact sequence. Another issue with the single-step decomposition strategy is that the predicted motion has zero velocity at the beginning and the end of each step. We can include the initial and goal root velocity in the task definition, but the task space and hence the required number of samples will increase. While random sampling is used in this work to generate the tasks, active learning~\cite{settles2009active} can reduce the required number of samples. 

Finally, we plan to extend the approach to include multi-contact locomotion with both hands and legs as potential contacts, and locomotion with varying contacts' heights (e.g. climbing stairs or uneven terrains). 


\IEEEtriggeratref{15}

\bibliographystyle{IEEEtran}
\bibliography{IEEEabrv,IEEEconf,main}

\end{document}